\documentclass[conference]{IEEEtran}
\IEEEoverridecommandlockouts

\usepackage[T1]{fontenc}
\usepackage{lmodern}
\usepackage{times}

\usepackage{amsmath,graphicx,hyperref,amssymb}
\usepackage{url}            
\usepackage{booktabs}       
\usepackage{amsfonts}       
\usepackage{nicefrac}       
\usepackage{microtype}      
\usepackage{titletoc}
\usepackage{pdfpages}
\usepackage{setspace}
\usepackage{subcaption}
\usepackage{graphicx}
\usepackage{amsmath}
\usepackage{multirow}
\usepackage{color}
\usepackage{colortbl}
\usepackage{xcolor}
\usepackage{setspace}
\usepackage{makecell}
\usepackage{cleveref}
\usepackage{algorithm}
\usepackage{balance}
\usepackage{algorithmicx}
\usepackage{algpseudocode}

\definecolor{lightkeycolor}{RGB}{255,240,240}
\definecolor{lightgray}{RGB}{234,234,234}

\begin{document}
\title{Diffusion-Guided Semantic Consistency for Multimodal Heterogeneity}
\author{
\IEEEauthorblockN{Jing Liu\IEEEauthorrefmark{1}\textsuperscript{,}\IEEEauthorrefmark{2}\textsuperscript{,}\IEEEauthorrefmark{3},
Zhengliang Guo\IEEEauthorrefmark{2},
Yan Wang\IEEEauthorrefmark{4},
Xiaoguang Zhu\IEEEauthorrefmark{5},
Yao Du\IEEEauthorrefmark{1},
Zehua Wang\IEEEauthorrefmark{1}\IEEEauthorrefmark{6},
Victor C. M. Leung\IEEEauthorrefmark{1}\textsuperscript{,}\IEEEauthorrefmark{7}\textsuperscript{,}\IEEEauthorrefmark{8}}
\IEEEauthorblockA{\IEEEauthorrefmark{1}The University of British Columbia,
\IEEEauthorrefmark{2}Fudan University,
\IEEEauthorrefmark{3}Duke Kunshan University,
\IEEEauthorrefmark{4}East China Normal University,\\
\IEEEauthorrefmark{5}University of California, Davis,
\IEEEauthorrefmark{6}China University of Mining and Technology,
\IEEEauthorrefmark{7}SMBU,
\IEEEauthorrefmark{8}Shenzhen University}
\thanks{This work is supported in part by the Guangdong Pearl River Talents Recruitment Program under Grant No. 2019ZT08X603, the Guangdong Pearl Rivers Talent Plan under Grant No. 2019JC01X235, the National Natural Science Foundation ofChina (No. 62406075), the National Key Research and Development Program of China (2023YFC3604802), the Shanghai Key Technology R\&D Program (Grant No. 25511107200), the Natural Sciences and Engineering Research Council (NSERC) of Canada under Grants Nos. RGPIN-2021-02970 and DGECR-202100187, and the Mitacs Project under Grant Nos. IT44479 and QJLI GR037230. ({Corresponding author: Yao Du}).}
}
\maketitle

\begin{abstract}
Federated learning (FL) is severely challenged by non-independent and identically distributed (non-IID) client data, a problem that degrades global model performance, especially in multimodal perception settings. Conventional methods often fail to address the underlying semantic discrepancies between clients, leading to suboptimal performance for multimedia systems requiring robust perception. To overcome this, we introduce SemanticFL, a novel framework that leverages the rich semantic representations of pre-trained diffusion models to provide privacy-preserving guidance for local training. Our approach leverages multi-layer semantic representations from the pre-trained Stable Diffusion model (including VAE-encoded latents and U-Net hierarchical features) to create a shared semantic space that aligns heterogeneous clients, facilitated by an efficient client-server architecture that offloads heavy computation to the server. A unified consistency mechanism, employing cross-modal contrastive learning, further stabilizes convergence. We conduct extensive experiments on benchmarks including CIFAR-10, CIFAR-100, and TinyImageNet under diverse heterogeneity scenarios. Our results demonstrate that SemanticFL surpasses existing federated learning approaches, achieving accuracy gains of up to 5.49\% over FedAvg, validating its effectiveness in learning robust models from heterogeneous and multimodal data for perception tasks.
\end{abstract}


\section{Introduction}
\label{sec:intro}
Federated learning (FL) enables collaborative model training on decentralized data, a critical capability in privacy-sensitive domains such as healthcare, autonomous vehicles, and embodied robotics~\cite{Karimireddy2019SCAFFOLDSC,du2023accelerating}. As multimedia and embodied AI systems increasingly rely on multimodal sensor fusion for robust perception~\cite{liu2025aligning}, the practical deployment of FL becomes fundamentally challenged by data heterogeneity, where clients possess non-independent and identically distributed (non-IID) data~\cite{liu2024fedgca}. In multimodal perception settings, this challenge becomes more severe as clients may hold varying combinations of data modalities, including images and text, with different quality and completeness~\cite{yoon2025vqfeddiff,liu2022learning}.

A primary technical barrier stemming from data heterogeneity is client drift, where local models diverge due to differing client data distributions~\cite{chen2025fedbip}. Conventional mitigation strategies enforce parameter-level consistency but often overlook underlying semantic discrepancies~\cite{yuan2024decentralized}. Consequently, local models overfit to their biased data distributions, compromising the generalization capability of the aggregated global model~\cite{Ma2022LayerwisedMA}. For instance, under a non-IID with a small concentration parameter, clients may possess data from only a few classes, making it difficult to learn a globally robust model~\cite{du2026decentralized}.

Recent approaches to mitigating client drift in federated learning have primarily focused on regularization and representation alignment. Methods like FedProx \cite{Li2020FedProx} add proximal terms to restrict local updates, while SCAFFOLD \cite{Karimireddy2019SCAFFOLDSC} uses control variates to correct client drift. Representation-based approaches such as MOON \cite{Li2021ModelContrastiveFL} employ model-level contrastive learning, and FedProto \cite{Tan2021FedProtoFP} learns global class prototypes for regularization. However, these methods operate within the limited parameter or feature space of client models themselves, constraining their ability to capture rich semantic relationships across heterogeneous data distributions~\cite{wang2022systematic,wang2024mgr3}.

In multimodal FL settings, contrastive learning has shown promise for aligning representations across missing modalities and heterogeneous feature spaces \cite{Yu2023MultimodalFL, Seo2024RelaxedCL}. Simultaneously, the integration of generative models into FL has gained traction, with recent works utilizing diffusion models for data augmentation \cite{wang2025feddifrc, yoon2025vqfeddiff} and knowledge distillation \cite{Najafi2024EnhancingGM}. While these generation-based approaches can be computationally expensive, they highlight the untapped potential of leveraging powerful generative models' intermediate representations for semantic guidance rather than data generation~\cite{liu2025anomaly,liu2025survey}.

The success of diffusion models in generative tasks has revealed their capacity to learn powerful, fine-grained semantic representations that capture complex data distributions and cross-modal correlations~\cite{gu2024federated}. Leveraging these rich representations offers a promising avenue to address the semantic heterogeneity challenge in FL~\cite{wang2025fedsc}. Although some studies have begun integrating diffusion models into FL for data augmentation or conditional generation \cite{wang2025feddifrc,yoon2025vqfeddiff}, their inherent representational power remains underexploited for direct semantic guidance in heterogeneous, multimodal settings~\cite{liu2026edgecloud,wu2025survey}. A comprehensive framework is therefore needed to harness these representations for semantic alignment without incurring prohibitive computational costs or violating privacy.

To address these challenges, we introduce SemanticFL, a novel framework that fundamentally rethinks how diffusion models can guide federated learning under severe data heterogeneity~\cite{liu2025multimodala,liu2026enhancing}. Rather than generating synthetic data or performing expensive on-client inference, our approach extracts rich semantic representations offline from a frozen Stable Diffusion model, creating a shared semantic space that serves as a stable anchor for heterogeneous local training. The framework operates through a three-stage pipeline: First, the server performs one-time offline extraction of multi-layer visual features from the diffusion U-Net architecture and textual embeddings from the CLIP encoder, capturing both fine-grained semantic patterns and high-level class concepts. Second, during federated training rounds, the server broadcasts compact pre-computed features alongside model parameters, enabling resource-constrained clients to benefit from powerful generative priors without local diffusion inference. Third, clients optimize their lightweight models through a unified objective that synergistically combines supervised classification, feature-level knowledge distillation from visual representations, and cross-modal contrastive learning between local features and textual semantics. 
By grounding local updates in a unified multimodal representation space, our design simultaneously addresses computational constraints and establishes semantic consistency across clients, thereby mitigating client drift while preserving privacy through feature-level rather than raw data collaboration.
The contributions are summarized as follows:
\begin{itemize}
    \item We propose a novel FL framework leveraging fine-grained diffusion representations for direct semantic guidance to address multimodal data heterogeneity.
    \item Our method introduces an efficient client-server architecture that decouples diffusion feature extraction from local training, enabling powerful semantic guidance for resource-constrained clients.
    \item Experimental validation on benchmark datasets under challenging non-IID settings demonstrates substantial performance improvements over existing approaches.
\end{itemize}

\begin{figure*}[t]
    \centering
    \includegraphics[width=\textwidth]{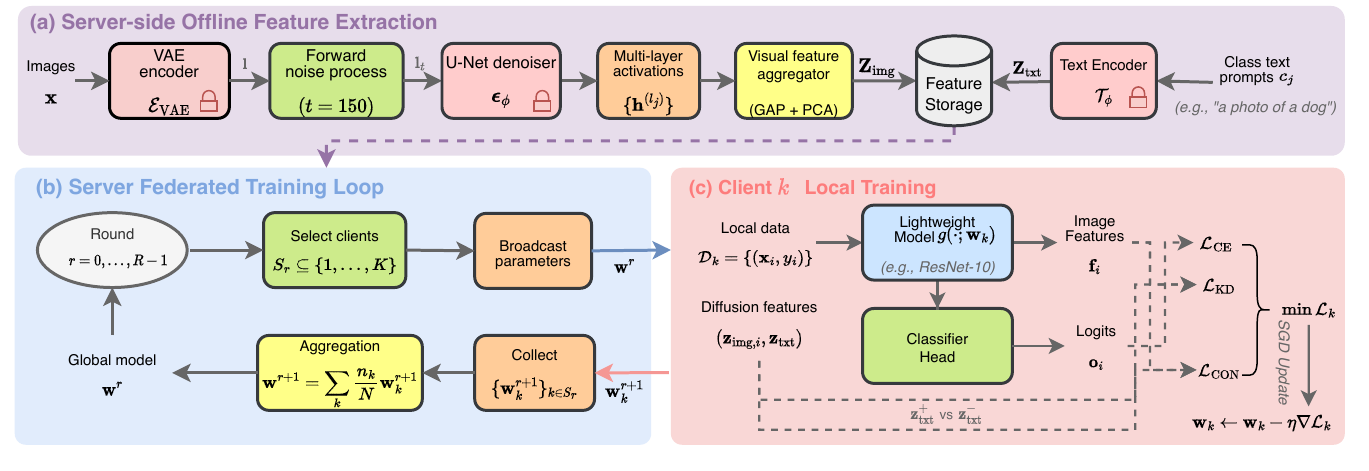}
    \caption{Overview of the SemanticFL framework. More details on components and symbols are described in Sec.~\ref{sec:method}.
    }
    \label{fig:framework_overview}
    \vspace{-15px}
\end{figure*}

\section{Related Work}
\label{sec:related}

\noindent\textbf{Federated Learning for Data Heterogeneity.}
Client drift caused by non-IID data remains a central challenge in federated learning. Regularization-based methods address this by constraining local updates: FedProx \cite{Li2020FedProx} adds a proximal term $\frac{\mu}{2}\|\mathbf{w}_k - \mathbf{w}^r\|^2$ to the local loss, whereas SCAFFOLD \cite{Karimireddy2019SCAFFOLDSC} employs control variates for drift correction. Representation alignment offers another avenue, with MOON \cite{Li2021ModelContrastiveFL} using model-level contrastive learning and FedProto \cite{Tan2021FedProtoFP} learning global class prototypes. Aggregation-level modifications include FedDisco \cite{Ye2023FedDiscoFL}, adjusting weights based on class distribution discrepancies, and layer-wise strategies for personalization \cite{Ma2022LayerwisedMA}. Moreover, knowledge distillation aligns client models with global consensus \cite{Dong2022FederatedCL}, while classifier post-calibration corrects bias \cite{Yu2023MultimodalFL}. However, operating within client model feature spaces limits these approaches, motivating our use of external semantic guidance.

\noindent\textbf{Multimodal and Contrastive Federated Learning.}
Multimodal federated learning faces challenges including missing modalities and heterogeneous feature space alignment~\cite{feng2025federated}. Contrastive learning enforces representation consistency to address these issues~\cite{kong2025federated}. Methods like FedRCL \cite{Seo2024RelaxedCL} employ supervised contrastive objectives with relaxed losses to prevent collapse and enhance transferability. CreamFL \cite{Yu2023MultimodalFL} uses inter-modal and intra-modal contrastive ensembles to regularize training and compensate for absent modalities. Despite their effectiveness in aligning standard client representations, these methods lack access to rich external semantic knowledge, which our pre-trained diffusion model provides for robust alignment across deeply heterogeneous clients.

\noindent\textbf{Diffusion Models in Federated Learning.}
Generative models have gained traction for combating data heterogeneity in federated learning. Early approaches used VAEs and GANs to synthesize data for local augmentation or server-side proxy datasets for distillation \cite{wang2025feddifrc}. Diffusion models now offer powerful alternatives: some generate synthetic data to augment local datasets \cite{ yoon2025vqfeddiff}, while others serve as data-free generators for knowledge distillation \cite{Najafi2024EnhancingGM}. Frameworks like Phoenix \cite{Jothiraj2023PhoenixAF} train diffusion models federatedly for privacy-preserving generation. Nevertheless, generation-based approaches incur computational costs and may not directly address semantic drift. Our work instead leverages pre-computed intermediate representations from diffusion models as direct semantic guidance, providing efficient and direct semantic regularization.

\section{Method}
\label{sec:method}

\subsection{Method Overview and Problem Formulation}
\label{sec:overview}
In federated learning, $K$ clients collaboratively train a global model with parameters $\mathbf{w}$ without sharing local data. Each client $k$ possesses a private dataset $\mathcal{D}_k = \{(\mathbf{x}_i, y_i)\}_{i=1}^{n_k}$ with $n_k$ samples, where the federated objective minimizes the weighted global loss $F(\mathbf{w}) = \sum_{k=1}^{K} \frac{n_k}{N} F_k(\mathbf{w})$, with $N = \sum_{k=1}^{K} n_k$ and $F_k(\mathbf{w}) = \mathbb{E}_{(\mathbf{x}, y) \sim \mathcal{D}_k} [\ell(g(\mathbf{x}; \mathbf{w}), y)]$ representing the local empirical risk. Under severe data heterogeneity, non-IID client distributions cause divergent local optima and ``client drift'' that degrades both convergence and generalization.

As illustrated in Fig.~\ref{fig:framework_overview}, SemanticFL addresses heterogeneity through diffusion-guided semantic alignment across three coordinated components. \textit{First}, during offline initialization, the server extracts semantic features from a frozen Stable Diffusion model~\cite{rombach2022highresolution}: visual representations $\mathbf{Z}_{\text{img}} = \{\mathbf{z}_{\text{img}, i}\}$ are computed via VAE encoding $\mathcal{E}_{\text{VAE}}$ (Eq.~\ref{eq:vae_encode}), controlled noising at timestep $t=150$ (Eq.~\ref{eq:forward_process_sample}), and multi-layer U-Net feature aggregation $\boldsymbol{\epsilon}_\phi$ (Eq.~\ref{eq:visual_feature_extraction}), while textual embeddings $\mathbf{Z}_{\text{txt}} = \{\mathbf{z}_{\text{txt}}^{(c)}\}_{c=1}^C$ are generated from class prompts via CLIP encoder $\mathcal{T}_\phi$. \textit{Second}, in each communication round $r$, the server broadcasts global parameters $\mathbf{w}^r$ alongside relevant diffusion features to a selected client subset $S_r \subseteq \{1, \ldots, K\}$. \textit{Third}, each client $k \in S_r$ performs local optimization by computing features $\mathbf{f}_i = g(\mathbf{x}_i; \mathbf{w}_k)$ and minimizing a unified objective $\mathcal{L}_k$ (Eq.~\ref{eq:final_loss}) that combines cross-entropy classification $\mathcal{L}_{\text{CE}}$, knowledge distillation from visual features $\mathcal{L}_{\text{KD}}$, and cross-modal contrastive alignment $\mathcal{L}_{\text{CON}}$ between local features and textual semantics. After local training, updated parameters $\mathbf{w}_k^{r+1}$ are aggregated via federated averaging (Eq.~\ref{eq:fed_agg}) to produce $\mathbf{w}^{r+1}$, where the entire architecture strategically decouples expensive diffusion inference from client-side training while establishing a shared semantic space for consistent alignment.

\subsection{Adaptive Multimodal Diffusion Representation}
Robust semantic guidance is provided by features extracted offline from a pre-trained Stable Diffusion v1.5 model~\cite{rombach2022highresolution}, which consists of three key components: a Variational Autoencoder (VAE) for image-latent conversion, a U-Net for denoising in latent space, and a CLIP-based text encoder for text conditioning.

\noindent\textbf{Visual Representation.} For each image $\mathbf{x}_i$ in the training set, we extract visual guidance features through the following pipeline. First, the image is encoded into the latent space using the VAE encoder:
\begin{equation}
\label{eq:vae_encode}
\mathbf{l}_i = \mathcal{E}_{\text{VAE}}(\mathbf{x}_i) \cdot \gamma,
\end{equation}
where $\gamma=0.18215$ is the standard SD scaling factor for numerical stability. Subsequently, we apply the forward diffusion process to add controlled noise at timestep $t=150$:
\begin{equation}
\label{eq:forward_process_sample}
\mathbf{l}_{i,t} = \sqrt{\bar{\alpha}_t}\mathbf{l}_i + \sqrt{1 - \bar{\alpha}_t}\boldsymbol{\epsilon}, \quad \boldsymbol{\epsilon} \sim \mathcal{N}(0, \mathbf{I}).
\end{equation}
Once obtained, the noisy latent representation $\mathbf{l}_{i,t}$ is fed into the U-Net denoising network $\boldsymbol{\epsilon}_\phi$, from which we extract intermediate activations from multiple hierarchical layers $\{l_j\}_{j=1}^J$. Letting $\mathbf{h}^{(l_j)}(\mathbf{l}_{i,t}, t)$ denote the feature map from layer $l_j$, these multi-scale features are then aggregated and projected to form the final visual representation:
\begin{equation}
\label{eq:visual_feature_extraction}
\mathbf{z}_{\text{img}, i} = \mathcal{P}\left(\text{Concat}[\{\text{GAP}(\mathbf{h}^{(l_j)}(\mathbf{l}_{i,t}, t))\}_{j=1}^J]\right),
\end{equation}
where GAP denotes global average pooling and $\mathcal{P}$ is a PCA-based dimensionality reduction to $d=512$ dimensions.

\noindent\textbf{Textual Representation.} For each class $j$ with text prompt $c_j$ (e.g., ``a photo of a dog''), the CLIP-based text encoder $\mathcal{T}_\phi$ in SD generates semantic embeddings $\mathbf{e}_j = \mathcal{T}_\phi(c_j)$. For class $y$, the positive feature is defined as $\mathbf{z}_{\text{txt}, y}^{+} = \mathbf{e}_y$, while negative features are $\mathbf{z}_{\text{txt}, y}^{-} = \{\mathbf{e}_j\}_{j \neq y}$, providing cross-modal semantic guidance for the contrastive loss in Sec. \ref{sec:consistency}.
\subsection{Efficient Client-Server Feature Extraction}
\label{sec:feature_extraction}
Large diffusion models incur substantial computational overhead, which poses a significant barrier for resource-constrained clients. To address this issue, we design an efficient client-server architecture that partitions the workload by offloading all heavy computations to the server.

\noindent\textbf{Server-Side Feature Generation.}
The server performs a one-time, offline extraction of all diffusion features before federated training begins, leveraging the complete Stable Diffusion v1.5 pipeline. For visual features, the process follows three stages:
\noindent\textit{Stage 1: Latent Encoding.} Each image $\mathbf{x}_i$ is first encoded into the latent space using the pre-trained VAE encoder (Eq.~\ref{eq:vae_encode}), thereby reducing the spatial resolution from $32 \times 32$ (CIFAR) to $4 \times 4$ in latent space while maintaining semantic information.
\noindent\textit{Stage 2: Controlled Noising.} The forward diffusion process (Eq.~\ref{eq:forward_process_sample}) adds Gaussian noise to the latent representation at timestep $t=150$, creating a partially noisy version $\mathbf{l}_{i,t}$ that retains sufficient semantic structure for feature extraction.
\noindent\textit{Stage 3: Multi-layer Feature Extraction.} The noisy latent $\mathbf{l}_{i,t}$ is passed through the U-Net denoising network, where intermediate activations from hierarchical layers (encoder, bottleneck, and decoder) are extracted using Eq.~\ref{eq:visual_feature_extraction} to capture multi-scale semantic information.
For textual features, class-specific prompts are processed through the CLIP text encoder $\mathcal{T}_\phi$ to generate $C$ semantic embeddings. All computed features $\mathbf{Z}_{\text{img}}$ and $\mathbf{Z}_{\text{txt}}$ are stored on the server for distribution to clients.

\noindent\textbf{Client-Side Lightweight Training.}
During each communication round, the server distributes only the compact, pre-computed features relevant to each client's local data~\cite{liu2026projecting}. Since feature extraction is decoupled from training, clients perform only the forward and backward passes of their smaller local models (e.g., ResNet-10~\cite{roy2021attentionbased}). Let $\mathbf{f}_i = g(\mathbf{x}_i; \mathbf{w}_k)$ denote the feature representation computed by client $k$'s model. The client then aligns $\mathbf{f}_i$ with the received diffusion features $\mathbf{z}_{\text{img}, i}$ and $\mathbf{z}_{\text{txt}}$ using the unified loss function (Eq.~\ref{eq:final_loss}). As a result, clients benefit from rich semantic guidance without incurring the computational cost of running the diffusion model locally.

\subsection{Unified Multimodal Consistency Mechanism}
\label{sec:consistency}
During local training on client $k$, model parameters $\mathbf{w}_k$ are optimized using a unified loss function $\mathcal{L}_k$ with three terms: classification loss $\mathcal{L}_{\text{CE}}$, knowledge distillation loss $\mathcal{L}_{\text{KD}}$, and cross-modal contrastive loss $\mathcal{L}_{\text{CON}}$.

For local data batch $\{(\mathbf{x}_i, y_i)\}_{i=1}^B$, the model produces logits $\mathbf{o}_i$ and L2-normalized features $\mathbf{f}_i$. The cross-entropy loss is:
\begin{equation}
\label{eq:ce_loss}
\mathcal{L}_{\text{CE}} = -\frac{1}{B} \sum_{i=1}^B \log\left(\frac{\exp(\mathbf{o}_{i, y_i})}{\sum_{c=1}^C \exp(\mathbf{o}_{i, c})}\right).
\end{equation}
The knowledge distillation loss aligns local features with diffusion representations:
\begin{equation}
\label{eq:kd_loss}
\mathcal{L}_{\text{KD}} = \frac{1}{B} \sum_{i=1}^B D_{\text{KL}}\left(\text{softmax}(\mathbf{z}_{\text{img}, i}) \parallel \text{softmax}(\mathbf{f}_i)\right).
\end{equation}
The cross-modal contrastive loss aligns visual features $\mathbf{f}_i$ with textual semantics using InfoNCE:
\begin{equation}
\label{eq:con_loss}
\mathcal{L}_{\text{CON}} = -\frac{1}{B} \sum_{i=1}^B \log \frac{\exp(s_i^+/\tau)}{\exp(s_i^+/\tau) + \sum_{j} \exp(s_{i,j}^-/\tau)},
\end{equation}
where $s_i^+ = \text{sim}(\mathbf{f}_i, \mathbf{z}_{\text{txt}, y_i}^{+})$ and $s_{i,j}^- = \text{sim}(\mathbf{f}_i, \mathbf{z}_j^{-})$.

\subsection{Federated Optimization and Training}
\label{sec:fed_opt}
The final objective for each client $k$ is to minimize the weighted sum of the three loss components over its local dataset $\mathcal{D}_k$:
\begin{equation}
\label{eq:final_loss}
\min_{\mathbf{w}_k} \mathcal{L}_k(\mathbf{w}_k) = \mathbb{E}_{(\mathbf{x}, y) \sim \mathcal{D}_k} \left[ \mathcal{L}_{\text{CE}} + \lambda_{\text{KD}}\mathcal{L}_{\text{KD}} + \lambda_{\text{CON}}\mathcal{L}_{\text{CON}} \right],
\end{equation}
where $\lambda_{\text{KD}}$ and $\lambda_{\text{CON}}$ are hyperparameters. After local training, the server aggregates the updated client models $\mathbf{w}_k^{r+1}$ from a subset of clients $S_r$ to produce the new global model $\mathbf{w}^{r+1}$ via federated averaging:
\begin{equation}
\label{eq:fed_agg}
\mathbf{w}^{r+1} \leftarrow \sum_{k \in S_r} \frac{n_k}{\sum_{j \in S_r} n_j} \mathbf{w}_k^{r+1}.
\end{equation}

\section{Experiments}
\label{sec:experiments}

\subsection{Experimental Setup}
\label{sec:experimental}

\noindent\textbf{Datasets and Heterogeneity Scenarios.}
Evaluation is conducted on several benchmark datasets. For standard classification, {CIFAR-10}, {CIFAR-100}~\cite{krizhevsky2010convolutional}, and {TinyImageNet}~\cite{le2015tiny} are used. Two heterogeneity scenarios are modeled: (1) {Label Shift}: Data is partitioned using a Dirichlet distribution with concentration parameter $\alpha \in \{0.05, 0.2, 0.5\}$, plus an extreme case (NID2). (2) {Imbalance Shift}: Long-tailed distributions are created with imbalance ratios $\rho \in \{10, 50, 100\}$.

\noindent\textbf{Implementation Details.}
All experiments are implemented in PyTorch on an NVIDIA RTX 4090 GPU. Our FL setup consists of $K=10$ total clients over $R=100$ communication rounds, with 5 clients randomly selected per round. Local training runs for $E=10$ epochs using an SGD optimizer (batch size 64, momentum 0.9, learning rate 0.01). Weight decay is $10^{-5}$ for CIFAR-10 and $10^{-4}$ for other datasets. For network architecture, the backbone model is ResNet-10 for CIFAR datasets and MobileNetV2 for TinyImageNet. For the proposed method, we leverage pre-trained Stable Diffusion v1.5~\cite{rombach2022highresolution} to extract multi-layer visual features from noisy latents ($t=150$) via denoising U-Net and class-specific textual embeddings via CLIP encoder, serving as semantic anchors for heterogeneity mitigation. The diffusion timestep is fixed at $t=150$, the feature dimension is $d=512$, the temperature is $\tau=0.05$, and loss balance hyperparameters are $\lambda_{\text{KD}}=1.0$ and $\lambda_{\text{CON}}=0.01$. Evaluation metric is top-1 accuracy. 

\begin{table*}[t]
\centering
\caption{Comparison of top-1 accuracy (\%) on CIFAR-10, CIFAR-100, and TinyImageNet across various non-IID scenarios.
}
\label{tab:main}
\setlength{\tabcolsep}{3.7mm}{
\resizebox{\textwidth}{!}{%
\begin{tabular}{l ccc ccc ccc}
\toprule
\multirow{2}{*}{\textbf{Method}} & \multicolumn{3}{c}{\textbf{CIFAR-10}} & \multicolumn{3}{c}{\textbf{CIFAR-100}} & \multicolumn{3}{c}{\textbf{TinyImageNet}} \\
\cmidrule(lr){2-4} \cmidrule(lr){5-7} \cmidrule(lr){8-10} & NID1$_{0.05}$ & NID1$_{0.2}$ & NID2 & NID1$_{0.05}$ & NID1$_{0.2}$ & NID2 & NID1$_{0.05}$ & NID1$_{0.2}$ & NID2 \\
\midrule
FedAvg~\cite{McMahan2016CommunicationEfficientLO} & 78.27 & 84.65 & 72.60 & 55.97 & 60.08 & 50.56 & 40.41 & 42.84 & 35.15 \\
FedProx~\cite{Li2020FedProx} & 78.42 & 84.59 & 72.81 & 56.27 & 60.21 & 50.29 & 40.20 & 42.16 & 35.62 \\
MOON~\cite{Li2021ModelContrastiveFL} & 80.79 & 86.10 & 73.35 & 56.79 & 61.48 & 51.81 & 40.79 & 43.63 & 36.11 \\
FedDisco~\cite{Ye2023FedDiscoFL} & 81.35 & 86.63 & 74.72 & 57.34 & 61.79 & 52.05 & 40.83 & 43.92 & 37.69 \\
FedCDA~\cite{Lanza2025JointCA} & 81.70 & 86.89 & 75.07 & 57.76 & 62.17 & 52.35 & 41.18 & 44.09 & 38.24 \\
FedRCL~\cite{Seo2024RelaxedCL} & 82.02 & 87.11 & 75.13 & 57.98 & 62.05 & 52.49 & 41.33 & 44.21 & 38.86 \\
FedDifRC~\cite{wang2025feddifrc} & 83.14 & 88.27 & 76.45 & 59.22 & 63.32 & 53.87 & 42.63 & 45.28 & 40.05 \\
\midrule
\textit{SemanticFL} & \textbf{83.76} & \textbf{88.94} & \textbf{77.12} & \textbf{59.85} & \textbf{64.05} & \textbf{54.58} & \textbf{43.25} & \textbf{45.93} & \textbf{40.89} \\
\bottomrule
\end{tabular}%
}}
\vspace{-15pt}
\end{table*}

\begin{table}[t]
\centering
\caption{Ablation study of components on CIFAR-10.
}
\label{tab:ablation}
\resizebox{\columnwidth}{!}{%
\begin{tabular}{cc ccc c}
\toprule
\textbf{AMDA} & \textbf{SFE} & \textbf{NID1$_{0.05}$} & \textbf{NID1$_{0.2}$} & \textbf{NID2} & \textbf{Average} \\
\midrule
& & 77.91 & 84.12 & 73.08 & 78.37 \\
\checkmark & & 82.15 & 85.67 & 76.21 & 81.34 \\
& \checkmark & 81.02 & 85.93 & 74.85 & 80.60 \\
\checkmark & \checkmark & \textbf{83.76} & \textbf{88.94} & \textbf{77.12} & \textbf{83.27} \\
\bottomrule
\end{tabular}%
}
\end{table}

\subsection{Results}
\label{sec:results}

\noindent\textbf{Main Results.}
Performance comparison between SemanticFL and state-of-the-art FL baselines is presented in Table \ref{tab:main}. We evaluate against comprehensive baselines including standard methods FedAvg~\cite{McMahan2016CommunicationEfficientLO} and FedProx~\cite{Li2020FedProx}, contrastive-based approaches MOON~\cite{Li2021ModelContrastiveFL} and FedRCL~\cite{Seo2024RelaxedCL}, aggregation-based methods FedDisco~\cite{Ye2023FedDiscoFL} and FedCDA~\cite{Lanza2025JointCA}, and recent diffusion-based approach FedDifRC~\cite{wang2025feddifrc}. Across three datasets and under challenging non-IID settings, SemanticFL consistently achieves superior accuracy. For instance, on CIFAR-10 with moderate heterogeneity ($\alpha=0.2$), it achieves 88.94\% accuracy, outperforming the next best method, FedDifRC~\cite{wang2025feddifrc}, by 0.67\%. Moreover, the performance gains are even more pronounced in highly heterogeneous settings; on CIFAR-100 under the extreme NID2 scenario, it achieves 54.58\%, a improvement of 0.71\% over FedDifRC~\cite{wang2025feddifrc}. Overall, these results validate the effectiveness of leveraging diffusion-based semantic guidance to mitigate client drift.

\noindent\textbf{Ablation Study.}
An ablation study, presented in Table \ref{tab:ablation}, dissects the contribution of each core component. Initially, the baseline model (FedAvg) achieves an average accuracy of 78.37\% on CIFAR-10, while integrating Adaptive Multimodal Diffusion Aggregation (AMDA) alone improves accuracy to 81.34\% (+2.97\%) and Server-side Feature Extraction (SFE) alone yields 80.60\% (+2.23\%). Notably, the complete framework combining both components achieves an average accuracy of 83.27\%, demonstrating a strong synergistic effect between the diffusion-based semantic guidance and the efficient feature extraction mechanism.

\begin{figure}[t]
\centering
\includegraphics[width=\columnwidth]{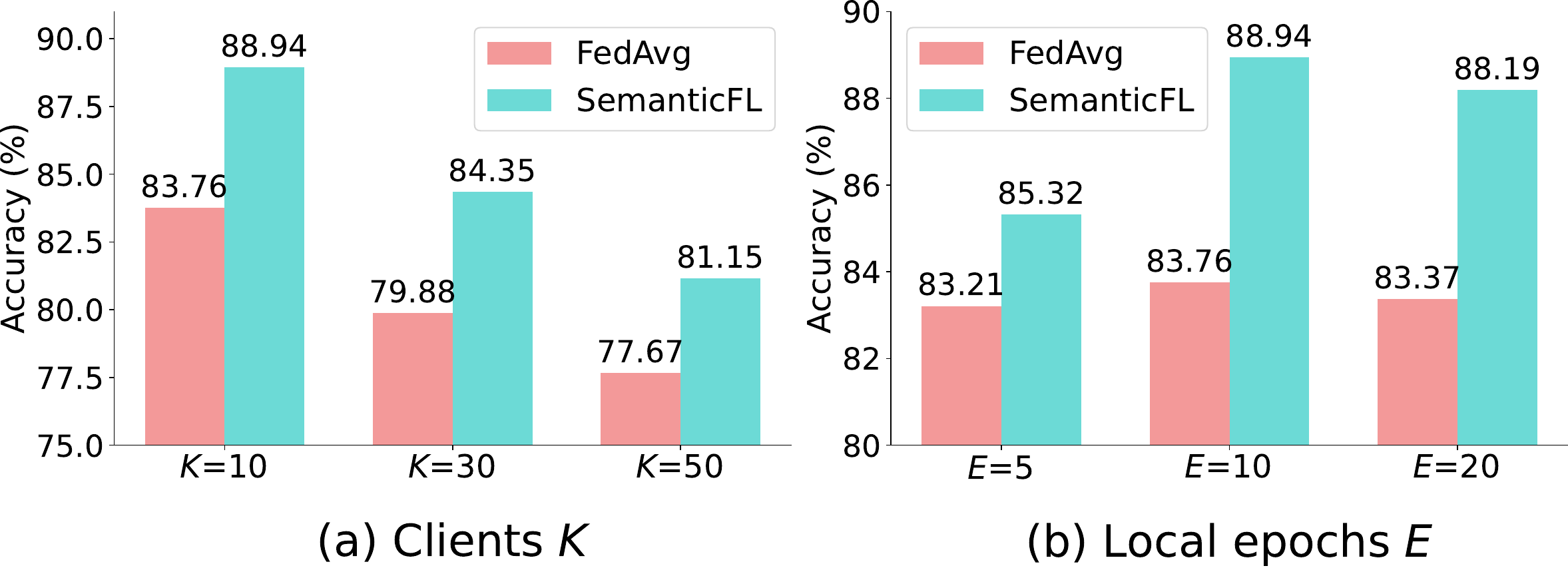}
\caption{Results of client number $K$ and local epochs $E$.
}
\label{fig:sensitivity}
\vspace{-15pt}
\end{figure}

\noindent\textbf{Parameter Analysis.}
The impact of key federated learning parameters on SemanticFL is analyzed in Fig. \ref{fig:sensitivity}, where the Fig. \ref{fig:sensitivity}(a) examines client number $K \in \{10, 30, 50\}$, showing that performance degrades with increasing clients due to heightened heterogeneity. Fig. \ref{fig:sensitivity}(b) studies local epochs $E \in \{5, 10, 20\}$, demonstrating optimal performance at $E=10$, balancing local convergence and client drift. Notably, SemanticFL maintains consistent performance advantages across all parameter settings, with relative improvements ranging from 2.11\% to 4.47\%, indicating the robustness of our diffusion-based semantic guidance mechanism.

\begin{figure}[t]
\centering
\includegraphics[width=\columnwidth]{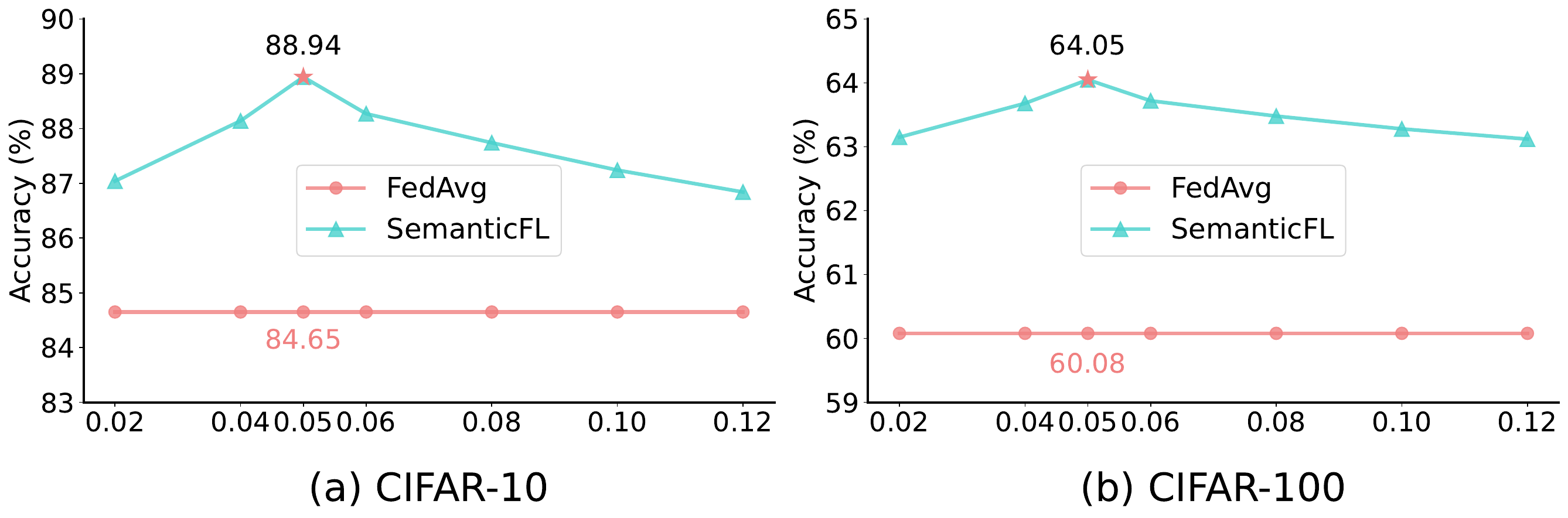}
\caption{Effect of temperature $\tau$ on CIFAR-10 and CIFAR-100.}
\label{fig:temperature}
\vspace{-15pt}
\end{figure}

\noindent\textbf{Temperature Sensitivity.}
We further investigate the sensitivity of contrastive temperature $\tau$ in Fig.~\ref{fig:temperature}. On CIFAR-10 (NID1$_{0.2}$), SemanticFL achieves peak accuracy of 88.94\% at $\tau=0.05$, with performance varying within a 2.10\% range across $\tau \in [0.02, 0.12]$, whereas on CIFAR-100 (NID1$_{0.2}$), the optimal accuracy of 64.05\% is similarly attained at $\tau=0.05$, with a smaller variation of 0.93\% due to the lower baseline accuracy. Importantly, SemanticFL consistently outperforms FedAvg (84.65\% on CIFAR-10, 60.08\% on CIFAR-100) across all tested temperature values, demonstrating robustness to hyperparameter selection and validating our default choice as both datasets exhibit similar trends where performance peaks at $\tau=0.05$ and gradually declines for smaller or larger values.

\begin{figure}[t]
\centering
\includegraphics[width=\columnwidth]{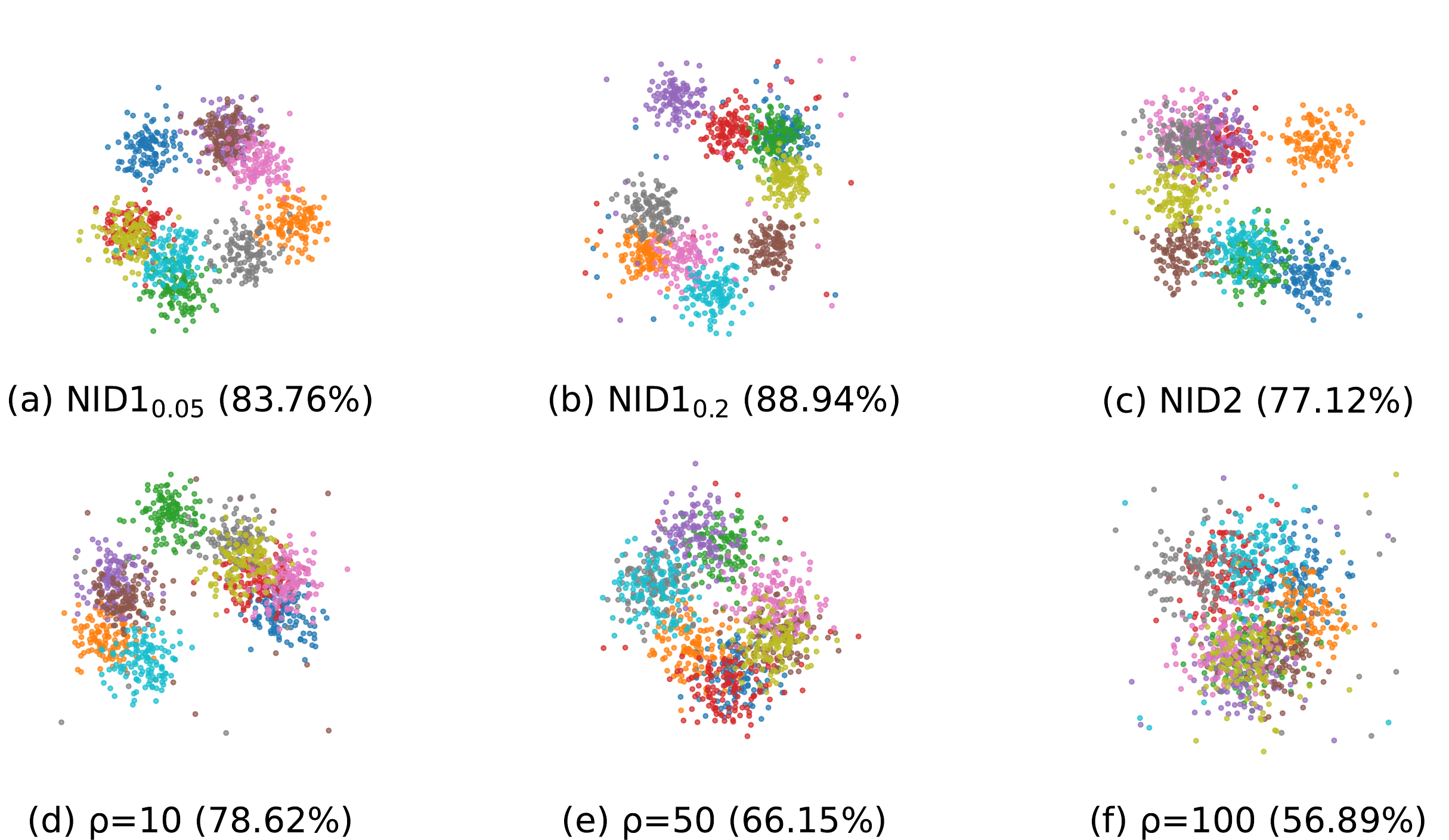}
\caption{t-SNE visualization of learned feature representations.
}
\vspace{-15px}
\label{fig:tsne}
\end{figure}

\noindent\textbf{Analysis of Learned Representations.}
Fig. \ref{fig:tsne} provides a t-SNE visualization of the feature representations learned on CIFAR-10, offering qualitative evidence for the quantitative results by showing that in settings with higher accuracy, such as NID1$_{0.2}$ (88.94\%), the feature clusters are more compact and well-separated. Even in challenging scenarios like the long-tailed setting with $\rho=100$ (56.89\%), SemanticFL still learns discriminative features albeit with more overlap, thereby supporting the claim that diffusion-based guidance helps the model learn a unified feature space resilient to data heterogeneity.

\noindent\textbf{Discussion.}
Collective results demonstrate that SemanticFL provides a robust solution to data heterogeneity in distributed multimodal perception. By leveraging semantic guidance from pre-trained diffusion models, it consistently achieves state-of-the-art performance across supervised and self-supervised scenarios. A limitation is the reliance on a powerful server for the initial offline feature extraction, which may not be available in all scenarios. Additionally, while key hyperparameters show robustness, their optimal values may vary across datasets and heterogeneity levels, warranting further investigation into adaptive tuning methods. 
Looking forward, the proposed semantic alignment mechanism has significant potential for embodied AI applications, where multi-systems must collaboratively process and understand diverse visual and textual content~\cite{feng2025multiagent}. In particular, the framework's ability to handle heterogeneous multimodal data makes it well-suited for distributed visual content analysis and interactive multimedia systems that require consistent semantic understanding across diverse sensing modalities.

\section{Conclusions}
\label{sec:conclusion}
In this paper, we introduce SemanticFL, a novel framework to mitigate data heterogeneity in federated learning by leveraging rich semantic guidance from pre-trained diffusion models. The approach demonstrates superior performance across three publicly available benchmarks under challenging non-IID conditions, opening new avenues for integrating large generative models into privacy-preserving distributed learning. Our semantic alignment mechanism shows particular promise for visual recognition and multimedia retrieval systems that require consistent cross-modal understanding across heterogeneous data sources. Future work will extend the framework to more complex multimedia modalities including video and audio content and investigate theoretical convergence guarantees under diverse heterogeneity scenarios.

\bibliographystyle{IEEEtran}
\begin{spacing}{0.99}
\bibliography{refs}
\end{spacing}
\end{document}